\documentclass[runningheads]{llncs}

\usepackage[utf8]{inputenc}
\usepackage[T1]{fontenc}
\usepackage[english]{babel}
\usepackage{amssymb}
\usepackage{multirow}
\usepackage{boldline}

\usepackage{graphicx, float}

\usepackage{mathtools}

\usepackage[nolist,nohyperlinks]{acronym}
\usepackage{cleveref}

\usepackage{subcaption}

\graphicspath{{images/}}

\title{Multi-Stage Bi-Atrial Segmentation Framework from 3D Late Gadolinium-Enhanced MRI using V-Net Family Models}

\titlerunning{Abbreviated paper title}
\author{
Hao Wen\inst{1}\orcidID{0000-0002-8856-0883}
\and
Jingsu Kang\inst{2}
}
\institute{
College of Science, China Agricultural University, Beijing, China\\
\email{wenh06@cau.edu.cn}, ~ \email{wenh06@gmail.com}
\and
Tianjin Medical University, Tianjin, China\\
\email{kangjingsu@tmu.edu.cn}
}

\begin{document}
\maketitle

\begin{abstract}
We report our multi-stage framework designed for the problem of multi-class bi-atrial segmentation from 3D late gadolinium-enhanced (LGE) MRI of the human heart. The pipeline consists of a preprocessing step using multidimensional contrast limited adaptive histogram equalization (MCLAHE); coarse region segmentation from MCLAHE-enhanced and down-sampled MRI using a V-Net family model; and fine segmentation from the coarse region using another V-Net model. Asymmetric loss is adopted to optimize the model weights.
\end{abstract}



\section{Introduction}
\label{sec:intro}


To improve the understanding of the human atrial anatomical structure, which is directly related to a wide range of widespread cardiac arrhythmia including atrial fibrillation, atrial flutter, atrial tachycardia, etc., a multi-class bi-atrial segmentation (MBAS) challenge is held on the CodaLab platform \cite{codalab_competitions_JMLR} as a part of the statistical atlases and computational modeling of the heart (STACOM) workshop in the 27th international conference on medical image computing and computer assisted intervention (MICCAI 2024).

This challenge extends the previous left atrium challenge \cite{Xiong_2021_mbas2021} and includes both the left and right atria and their walls. We propose a multi-stage framework to tackle this problem.

\section{Materials and Methods}
\label{sec:methods}


The challenge provides 70 3D LGE-MRI scans with mask labels for method development. The training data have a fixed spacing of 0.625, 0.625, and 2.5mm, typically with resolutions of $640 \times 640 \times 44$ and $576 \times 576 \times 44.$ We use all data in training the models without any data left out for model selection.

\begin{figure}
\centering
\includegraphics[width=0.8\linewidth]{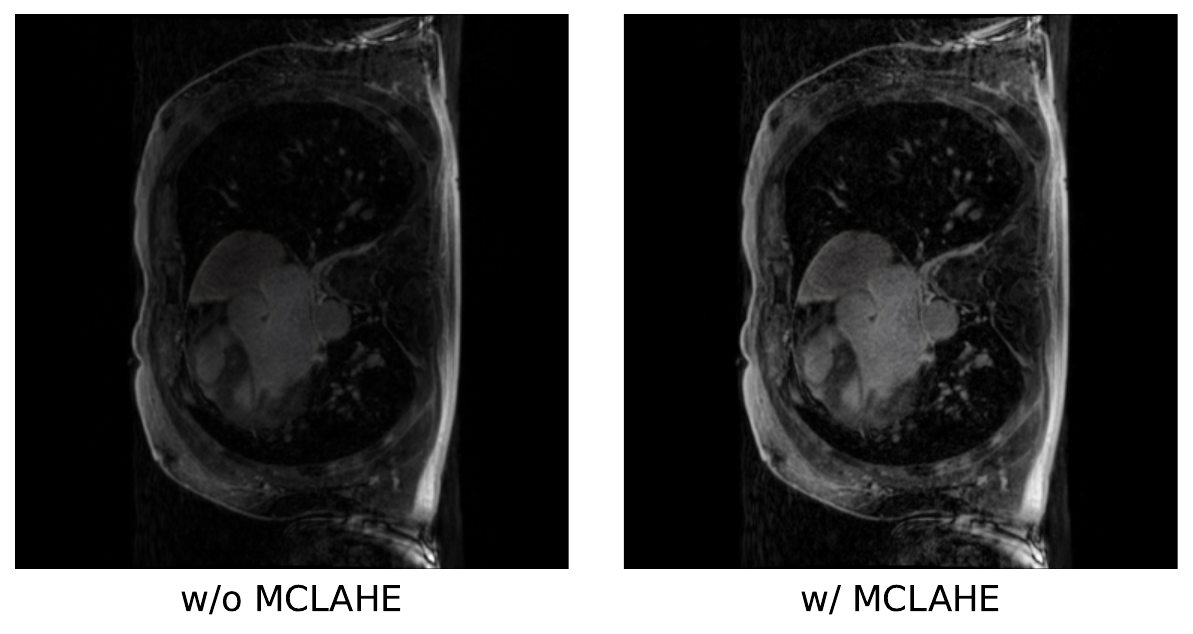}
\caption{An example (slice) of MCLAHE applied on MRI data. Left: raw data without MCLAHE applied; right: data with MCLAHE applied}
\label{fig:mclahe}
\end{figure}

\textbf{Pipeline step 1 - preprocessing}. Raw MRI scans are enhanced with multidimensional contrast limited adaptive histogram equalization (MCLAHE) \cite{Stimper_2019_mclahe} which is a quite useful technique in various medical image segmentation problems. A qualitative example of the effectiveness of MCLAHE is illustrated in \Cref{fig:mclahe}. The enhanced MRIs are rescaled by zero-padding and cropping so that each has a uniform resolution of $576 \times 576 \times 48.$

\textbf{Pipeline step 2 - coarse region detection}. The atrium occupies only a relatively small proportion of an MRI scan, so we split the segmentation into two stages. In the first stage, we detect a cubic region that contains the atrium, which we treat as a binary segmentation task. This task is accomplished with a neural network model chosen from the V-Net family \cite{Milletari_2016VNet,Zhou_2018_UNet++} on down-sampled (to resolution $144 \times 144 \times 48$) MRI data.

\textbf{Pipeline step 3 - fine multi-class segmentation}. We crop a region of size $256 \times 256 \times 48$ from the rescaled MRI data (obtained in step 1) based on the coarse atrium region (detected in step 2). This task is treated as a 4-class (left atrium, right atrium, the wall, and an additional background class) segmentation task. The segmentation model has a similar architecture to the coarse region detection model. The difference lies in the number of channels of the output layer.

\textbf{V-Net family models and training setups}. Due to the time limit, we conducted experiments on two types of V-Net models, namely the vanilla V-Net model \cite{Milletari_2016VNet} and a nested V-Net model (V-Net++) \cite{Zhou_2018_UNet++}. Since the coarse atrium region detection and fine multi-class segmentation steps are both highly imbalanced problems, we use the asymmetric loss \cite{ridnik2021asymmetric_loss} which has proven useful in such problems. The definition is given in \Cref{eq:asymmetric_loss} as follows:
\begin{equation}
\label{eq:asymmetric_loss}
L = -y \cdot (1-p)^{\gamma_{+}} \log(p) - (1-y) \cdot (p_m)^{\gamma_{-}} \log(1-p_m),
\end{equation}
where $y$ is the ground truth label, $p$ is the output probability of the model, $m$ is the probability margin, $p_m = \max(p - m, 0)$ is the shifted probability, and $\gamma_{+}$ and $\gamma_{-}$ are the positive and negative focusing parameters, respectively. The margin $m$ and the focusing parameters $\gamma_{+}$ and $\gamma_{-}$ are tunable hyperparameters. However, we did not do hyperparameter searching. Instead, we fixed $\gamma_{+} = 1,$ $\gamma_{-} = 4,$ and $m = 0.05$ in our experiments.

The \texttt{AdamW} optimizer \cite{adamw_amsgrad} is used, along with the \texttt{OneCycle} \cite{smith2019one_cycle} learning rate scheduler (initial learning rate: $0.0033,$ maximum learning rate: $0.0087$). The total number of training epochs is set to 100. No early stopping callback is set since we leave no data from the public training data for model selection.

\section{Experiments and Results}
\label{sec:results}

\begin{figure}[!htp]
\centering
\begin{subfigure}[t]{0.49\linewidth}
\centering
\includegraphics[width=\textwidth]{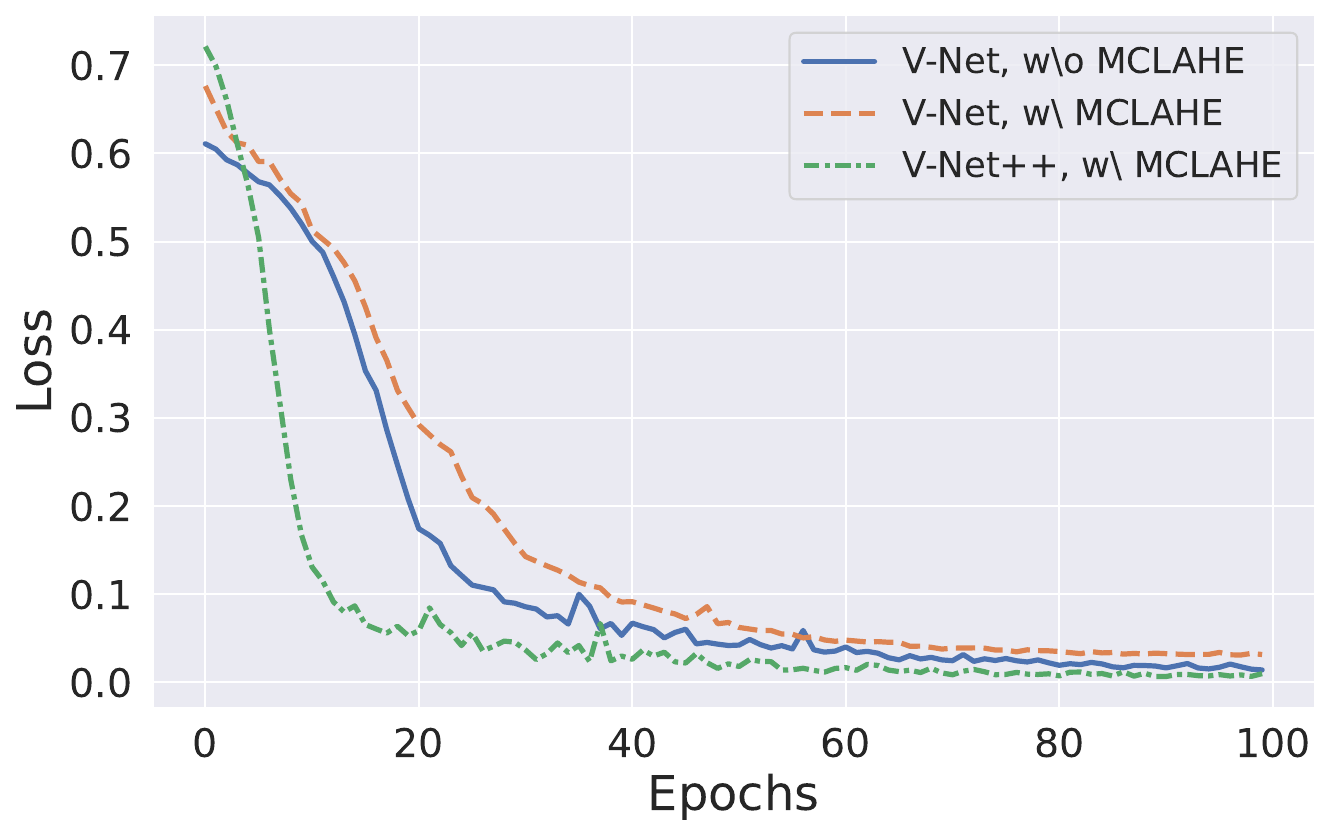}
\caption[]
{Train loss curves of the coarse atrium region detection models.}
\label{fig:coarse-detection-train-loss}
\end{subfigure}
\hfill
\begin{subfigure}[t]{0.49\linewidth}
\centering
\includegraphics[width=\textwidth]{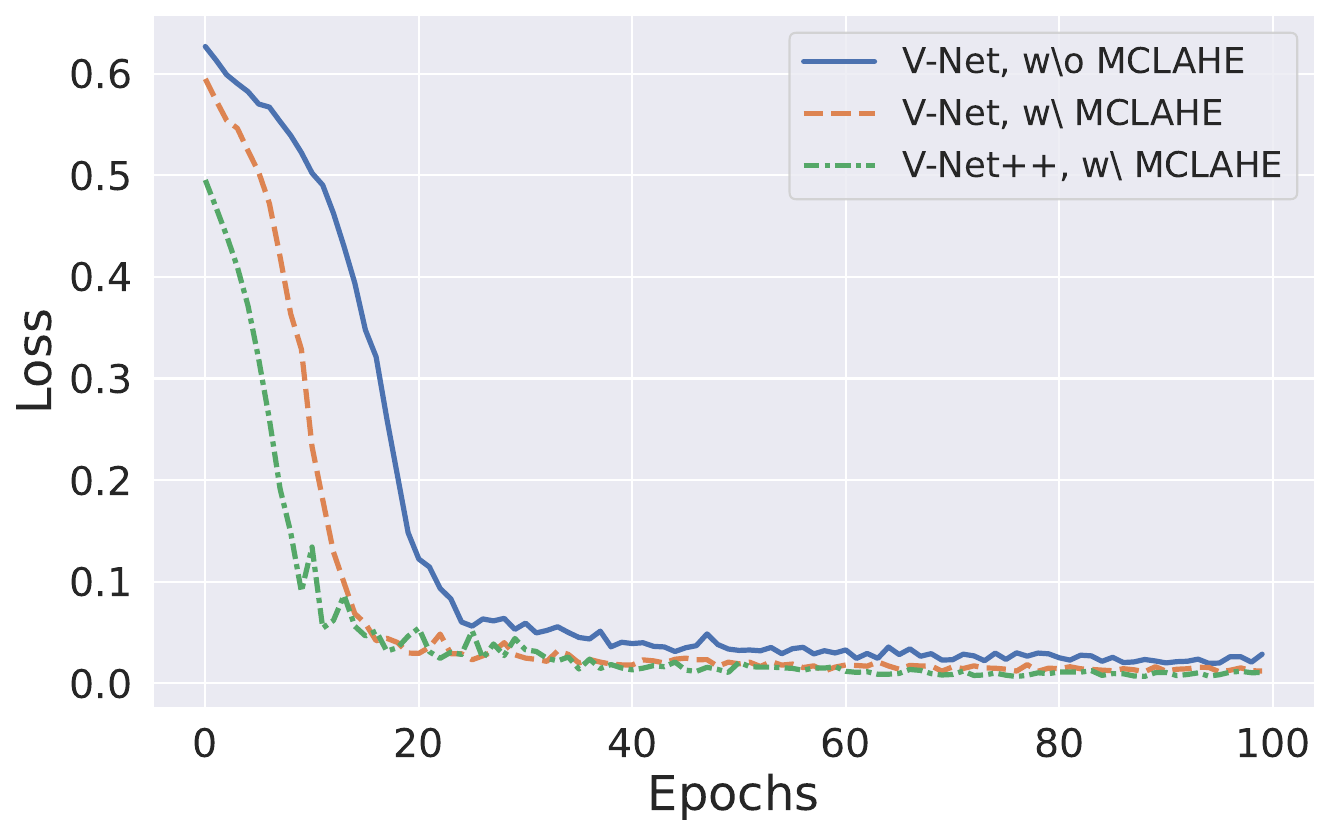}
\caption[]
{Train loss curves of the fine multi-class segmentation models.}
\label{fig:fine-segmentation-train-loss}
\end{subfigure}
\caption[]
{Loss curves of the models on the training data.}
\label{fig:train-loss-curves}
\end{figure}

We conduct 3 sets of experiments with different combinations of the model architecture and the usage of the MCLAHE enhancement preprocessing step:
\begin{enumerate}
\item V-Net, without MCLAHE applied;
\item V-Net, with MCLAHE applied;
\item nested V-Net (V-Net++), with MCLAHE applied.
\end{enumerate}
The loss curves of the above 3 experiments are collected in \Cref{fig:train-loss-curves}. Unsurprisingly, the V-Net++ models have the lowest loss and converge the fastest, since the V-Net++ models own a more complicated architecture, thus having better learning capability.


However, this is not the case when transferability is considered. The challenge allows 3 tries on a public validation set which consists of 30 3D LGE-MRI scans without mask labels. The above-mentioned trained models were tested on this validation set and the results are presented in \Cref{tab:val-results}. The vanilla V-Net model with an MCLAHE preprocessing step performed the best. Comparing the top two rows, a huge performance can be observed with MCLAHE.

\begin{table}[]
\centering
\setlength\tabcolsep{6pt}
\setlength\extrarowheight{1pt}
\begin{tabular}{@{\extracolsep{4pt}}rcllllll@{}}
\hlineB{3.5}
\multirow[c]{2}{*}{Model} & \multirow[c]{2}{*}{MCLAHE} & \multicolumn{2}{c}{Wall} & \multicolumn{2}{c}{Right Atrium} & \multicolumn{2}{c}{Left Atrium} \\ \cline{3-8}
 &  & Dice & HD95 & Dice & HD95 & Dice & HD95 \\
\hlineB{2.5}
V-Net & $\times$ & 46.2 & 7.07 & 82.82 & 16.14 & 84.96 & 10.29 \\
V-Net & $\checkmark$ & \bfseries 55.91 & \bfseries 5.10 & \bfseries 86.12 & \bfseries 6.77 & \bfseries 88.15 & \bfseries 5.81 \\
V-Net++ & $\checkmark$ & 55.11 & 6.16 & 81.40 & 7.98 & 87.43 & 6.78 \\
\hlineB{3.5}
\end{tabular}
\caption{Model performances on the validation set.}
\label{tab:val-results}
\end{table}

The metrics used for evaluating the models, or more precisely the whole pipeline, are
\begin{enumerate}
\item Dice: dice similarity coefficient (Dice);
\item HD95: 95th percentile of the symmetric Hausdorff distances between the region of interest (the wall, the left atrium, and the right atrium respectively) in two MRIs.
\end{enumerate}

The dice similarity coefficient is defined in \Cref{eq:dice}
\begin{equation}
\label{eq:dice}
Dice = \dfrac{2{\mathit{TP}}}{2{\mathit{TP}}+{\mathit{FP}}+{\mathit{FN}}},
\end{equation}
where the true positive (TP), false positive (FP), and false negative (FN) are computed in a pixel-wise manner. The dice similarity coefficient is positively correlated with the model performance.

For two regions $A, B$ in the Euclidean space, HD95 equals the 95th percentile of the set defined in \Cref{eq:hd95}
\begin{equation}
\label{eq:hd95}
\left\{ d(a, B) ~ : ~ a \in A \right\} \cup \left\{ d(b, A) ~ : ~ b \in B \right\},
\end{equation}
where $\displaystyle d(a, B) = \inf_{b\in B} d(a, b),$ $d(a, b)$ is the distance of two points in the Euclidean space. HD95 is more stable to outliers compared to the Hausdorff distance which is defined as
\begin{equation*}
HD = \max \left\{ \sup_{a\in A} d(a, B), \sup_{b\in B} d(b, A) \right\}.
\end{equation*}
HD95 is negatively correlated with model performance.

\section{Discussion and Conclusion}
\label{sec:discussion}


The results presented in \Cref{sec:results} demonstrate that the multi-stage framework proposed in this work is effective for the bi-atrial segmentation task. Some interesting phenomena related to the complexity and transferability of the models are also discovered in the numerical experiments. MCLAHE again proves its significant role in the field of medical image analysis.

There are limitations of this work. First, models with more advanced architectures are not tested, including more V-Net family models, transformer-based models \cite{Hatamizadeh_2022_UNETR} and the recently proposed ``segment anything'' (SA) framework \cite{Ma_2024_Segment}. Due to time and resource limitations, our numerical experiments are not thorough. Many of the important hyperparameters were set empirically without performing hyperparameter searching. The asymmetric loss function, which is distribution (cross-entropy) based, proves effective but not optimal. According to a previous comprehensive research work \cite{Ma_2021_LossOdyssey} on loss functions for medical image segmentation, compound loss functions (e.g. both distribution-based and region-based) are the most robust losses, especially for highly imbalanced segmentation tasks. This point has not been verified in this study and would be left as a future work.

\section*{Availability of code}
The code used in this study is publicly available at\\
\url{https://github.com/wenh06/MBAS2024}.

\section*{Acknowledgements}
The authors thank Professor Deren Han from Beihang University for helping accomplish this work.

\bibliographystyle{IEEEtran}

\bibliography{IEEEabrv,references}

\end{document}